\documentclass[10pt,twocolumn,letterpaper]{article}

\usepackage{cvpr}
\usepackage{times}
\usepackage{epsfig}
\usepackage{graphicx}
\usepackage{amsmath}
\usepackage{amssymb}

\usepackage{url}            
\usepackage{booktabs}       
\usepackage{amsfonts}       
\usepackage{nicefrac}       
\usepackage{microtype}      
\usepackage{textcomp}
\usepackage{xcolor}
\usepackage{multirow}

\usepackage{mathtools}
\usepackage{caption}
\usepackage{subcaption}
\usepackage{paralist}
\usepackage{array}
\usepackage{placeins}
\usepackage{epstopdf}
\usepackage[titletoc,title]{appendix}

\usepackage[ruled]{algorithm2e}
\makeatletter
\def\BState{\State\hskip-\ALG@thistlm}
\makeatother

\usepackage{color}
\usepackage{enumitem}
\setitemize{noitemsep,topsep=0pt,parsep=0pt,partopsep=0pt}

\usepackage{authblk}

\usepackage[pagebackref=true,breaklinks=true,letterpaper=true,colorlinks,bookmarks=false]{hyperref}

\cvprfinalcopy 

\def\cvprPaperID{} 

\ifcvprfinal\fi
\begin{document}

\title{Generating Classification Weights with GNN Denoising Autoencoders for Few-Shot Learning}

\author[1,2]{Spyros Gidaris}
\author[1]{Nikos Komodakis}
\affil[1]{University Paris-Est, LIGM, Ecole des Ponts ParisTech}
\affil[2]{valeo.ai}

\maketitle

\begin{abstract}
Given an initial recognition model already trained on a set of base classes, the goal of this work is to develop a meta-model for few-shot learning. The meta-model, given as input some novel classes with few training examples per class, must properly adapt the existing recognition model into a new model that can correctly classify in a unified way both the novel and the base classes. To accomplish this goal it must learn to output the appropriate classification weight vectors for those two types of classes. To build our meta-model we make use of two main innovations: we propose the use of a Denoising Autoencoder network (DAE) that (during training) takes as input a set of classification weights corrupted with Gaussian noise and learns to reconstruct the target-discriminative classification weights. In this case, the injected noise on the classification weights serves the role of regularizing the weight generating meta-model. Furthermore, in order to capture the co-dependencies between different classes in a given task instance of our meta-model, we propose to implement the DAE model as a Graph Neural Network (GNN). In order to verify the efficacy of our approach, we extensively evaluate it on ImageNet based few-shot benchmarks and we report strong results that surpass prior approaches. The code and models of our paper will be published on:
\url{https://github.com/gidariss/wDAE_GNN_FewShot}
\end{abstract}

\section{Introduction}

Over the last few years, deep learning has achieved impressive results on various visual understanding tasks,
such as image classification~\cite{krizhevsky2012imagenet, simonyan2014very, szegedy2015going, he2016deep},
object detection~\cite{ren2015faster}, or semantic segmentation~\cite{chen2018deeplab}.
However, their success heavily relies on the ability to apply gradient based optimization routines, which are computationally expensive, and 
having access to a large dataset of training data, which often is very difficult to acquire. 
For example, in the case of image classification, it is required to have available thousands or hundreds of training examples per class and the optimization routines consume hundreds of GPU days.
Moreover, the set of classes that a deep learning based model can recognize remains fixed after its training.
In case new classes need to be recognized, then it is typically required to collect thousands / hundreds of training examples for each of them, and re-train or fine-tune the model on those new classes.
Even worse, this latter training stage would lead the model to ``forget" the initial classes on which it was trained.
In contrast, humans can rapidly learn a novel visual concept from only one or a few examples and reliably recognize it later on. 
The ability of fast knowledge acquisition is assumed to be related with a meta-learning process in the human brain that exploits past experiences about the world when learning a new visual concept. 
Even more, humans do not forget past visual concepts when learning a new one.
Mimicking that behavior with machines is a challenging research problem, with many practical advantages, and is also the topic of this work. 

Research on this subject is usually termed \emph{few-shot object recognition}.
More specifically, few-shot object recognition methods tackle the problem of learning to recognize a set of classes given access to only a few training examples for each of them.
In order to compensate for the scarcity of training data they employ meta-learning strategies that learn how to efficiently recognize a set of classes with few training data by being trained on a distribution of such few-shot tasks (formed from the dataset available during training) that are similar (but not the same) to the few-shot tasks encountered at test time ~\cite{vinyals2016matching}. 
Few-shot learning is also related to transfer learning since the learned meta-models solve a new task by exploiting the knowledge previously acquired by solving a different set of similar tasks.
There is a broad class of few-shot learning approaches, including, among many, 
metric-learning-based approaches that learn a distance metric between a test example and the training examples~\cite{vinyals2016matching, snell2017prototypical, koch2015siamese, yang2018learning, wang2018low},
methods that learn to map a test example to a class label by accessing memory modules that store the training examples of that task~\cite{garcia2017few, mishra2018simple, santoro2016meta, kaiser2017learning, munkhdalai2017meta}, 
approaches that learn how to generate model parameters for the new classes given access to the few available training data of them~\cite{gidaris2018dynamic, qi2018low, gomez2005evolving, qiao2017few, ha2016hypernetworks}, 
gradient descent-based approaches~\cite{ravi2016optimization, finn2017model, andrychowicz2016learning} that learn how to rapidly adapt a model to a given few-shot recognition task via a small number of gradient descent iterations,
and training data hallucination methods~\cite{hariharan2017low, wang2018low} that learn how to hallucinate more examples of a class given access to its few training data.
 
\paragraph{Our approach.} 
In our work we are interested in learning a meta-model that is associated with a recognition model already trained on set of classes (these will be denoted as \emph{base classes} hereafter). Our goal is to train this meta-model so as to learn to adapt the above recognition model to a set of novel classes, for which there are only very few training data available (e.g., one or five examples), while at the same time maintaining the recognition performance on the base classes.
Note that, with few exceptions~\cite{hariharan2017low, wang2018low, gidaris2018dynamic, qi2018low, qiao2017few}, most prior work on few-shot learning neglects to fulfill the second requirement.
In order to accomplish this goal we follow the general paradigm of model parameter generation from few-data~\cite{gidaris2018dynamic, qi2018low, gomez2005evolving, qiao2017few}.
More specifically, we assume that the recognition model has two distinctive components,
a feature extractor network, which (given as input an image) computes a feature representation, and
a feature classifier, which (given as input the feature representation of an image) classifies it to one of the available classes by applying a set of classification weight vectors (one per class) to the input feature.
In this context, in order to be able to recognize novel classes one must be able to generate classification weight vectors for them. 
So, the goal of our work is to learn a meta-model that fulfills exactly this task: i.e., given a set of novel classes with few training examples for each of them, as well as the classification weights of the base classes, it learns to output a new set of classification weight vectors (both for the base and novel classes) that can then be used from the feature classifier in order to classify in a unified way both types of classes.

\paragraph{DAE based model parameters generation.} 
Learning to perform such a meta-learning task, i.e., inferring the classification weights of a set of classes, 
is a difficult meta-problem that requires plenty of training data in order to be reliably solved.
However, having access to such a large pool of data is not always possible; 
or otherwise stated the training data available for learning such meta-tasks might never be enough.
In order to overcome this issue we build our meta-model based on a Denoising Autoencoder network (DAE). During training, this DAE network takes as input a set of classification weights corrupted with additive Gaussian noise, and is trained to reconstruct the target-discriminative classification weights.
The injected noise on the inputs of our DAE-based parameter generation meta-model helps in the regularization of the learning procedure thus allowing us to avoid the danger of overfitting on the training data.
Furthermore, thanks to the theoretical interpretation of DAEs provided in~\cite{alain2014regularized}, our DAE-based meta-model is able to approximate gradients of the log of conditional distribution of the classification weights given the available training data by computing the difference between the input weights and the reconstructed weights~\cite{alain2014regularized}. 
Thus, starting by an initial (but not very accurate) estimate of classification weights, 
our meta-model is able to perform gradient ascents step(s) that will move the classification weights towards more likely configurations (when conditioned on the given training data).
In order to properly apply the DAE framework in the context of few-shot learning, 
we also adapt it so as to follow the episodic formulation~\cite{vinyals2016matching} typically used in few-shot learning. This further improves the performance of the parameters generation meta-task by forcing it to reconstruct more discriminative classification weights.

\paragraph{Building the model parameters DAE as a Graph Neural Network.}
\begin{figure}
\renewcommand{\figurename}{Figure}
\renewcommand{\captionlabelfont}{\bf}
\renewcommand{\captionfont}{\small} 
\begin{center}
\includegraphics[width=0.5\textwidth]{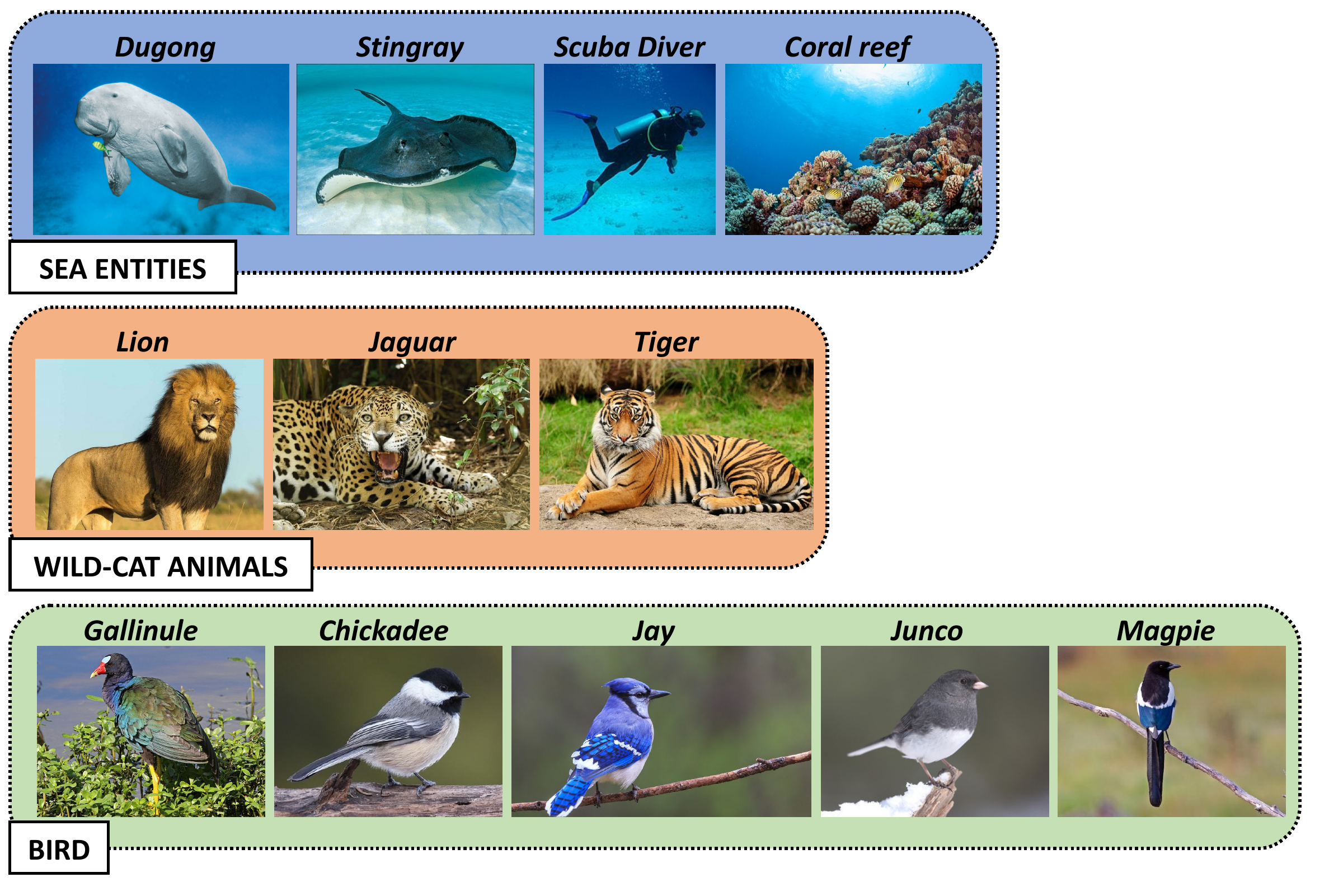}
\end{center}
\caption{
Some classes (\eg, wild-cat animals, birds, or sea creatures) are semantically or visually similar. Thus it is reasonable to assume that there are correlations between their classification weight vectors that could be exploited in order to reconstruct a more discriminative classification weight vector for each of them.}
\label{fig:visually_similar}
\end{figure}
Reconstructing classification weights conditioned only on a few training data, e.g., one training example per class, 
is an ill-defined task for which it is crucial to exploit as much of the available information on the training data of base and novel classes as possible.
In our context, 
one way to achieve that is by allowing the DAE model to learn the structure of the \emph{entire} set of classification weights that has to reconstruct on each instance (i.e., episode) of that task. 
We believe that such an approach is more principled and can reconstruct more distriminative classification weight vectors than reconstructing the classification weight of each class \emph{independently}. 
For example, 
considering that some of the classes (among the classes whose classification weight vectors must be reconstructed by the DAE model in a given task instance) are semantically or visually similar, 
such as different species of birds or see creatures (see Figure~\ref{fig:visually_similar}),
it would make sense to assume that there are correlations to their classification weight vectors that could be exploited in order to reconstruct a more discriminative classification weight vector for each of them.
In order to capture such co-dependencies between different classes (in a given task instance of our meta-model),
we implement the DAE model as a Graph Neural Network (GNN).
This is a family of deep learning networks designed to process an unordered set of entities (in our case a set of classes) associated with a graph $G$ such that they take into account their inter-entity relationships (in our case inter-class relationships) when making predictions about them~\cite{gori2005new,scarselli2009graph, li2015gated, duvenaud2015convolutional, sukhbaatar2016learning}
(for a recent survey on models and applications of deep learning on graphs see also Bronstein~\etal~\cite{bronstein2017geometric}).

Related to our work, Gidaris and Komodakis~\cite{gidaris2018dynamic} also tried to capture such class dependencies in the context of few-shot learning, 
by predicting the classification weight of each novel class as a mixture of the base classification weights through an attention mechanism. 
In contrast to them, we consider the dependencies that exist between all the classes, both novel and base (and not of a novel class with the base ones as in~\cite{gidaris2018dynamic}) and try to capture them in a more principled way through GNN architectures, which are more expressive compared to the simple attention mechanism proposed in~\cite{gidaris2018dynamic}.
Graph Neural Networks have been also used in the context of few-shot learning by Garcia and Juan~\cite{garcia2017few}. 
However, in their work, they give as input to the GNN the labeled training examples and the unlabeled test examples of a few-shot problem and they train it to predict the label of the test examples. 
Differently from them, in our work we provide as input to the GNN some initial estimates of the classification weights of the classes that we want to learn and we train them to reconstruct more discriminative classification weights.
Finally, Graph Neural Networks have been applied on a different but related problem, that of zero-shot learning~\cite{wang2018zero,kampffmeyer2018rethinking}, for regressing classification weights. 
However, in this line of work they apply the GNNs to knowledge graphs provided by external sources (e.g., word-entity hierarchies) while the input that is given to the GNN for a novel class is its word-embedding.  
Differently from them, in our formulation we do not consider any side-information (i.e., knowledge graphs or word-embeddings), which makes our approach more agnostic to the domain of problems that can solve and the existence of such knowledge graphs.  

To sum up, our contributions are, 
(1) the application of Denoising Autoencoders in the context of few-shot learning, 
(2) the use of Graph Neural Network architectures for the classification weight generation task, and 
(3) performing detailed experimental evaluation of our model on ImageNet-FS~\cite{hariharan2016low} and MiniImageNet~\cite{vinyals2016matching} datasets and achieving state-of-the-art results on ImageNet-FS, MiniImageNet, and \emph{tiered}-MiniImageNet~\cite{ren2018meta} datasets.

In the following sections, 
we describe our classification weights generation methodology in~\S\ref{sec:methodology},
we provide experimental results in~\S\ref{sec:experimentalresults}, and finally 
we conclude in~\S\ref{sec:conclusion}.

\section{Methodology} \label{sec:methodology}
\begin{figure*}[t]
\renewcommand{\figurename}{Figure}
\renewcommand{\captionlabelfont}{\bf}
\renewcommand{\captionfont}{\small} 
\begin{center}
\includegraphics[width=0.8\textwidth]{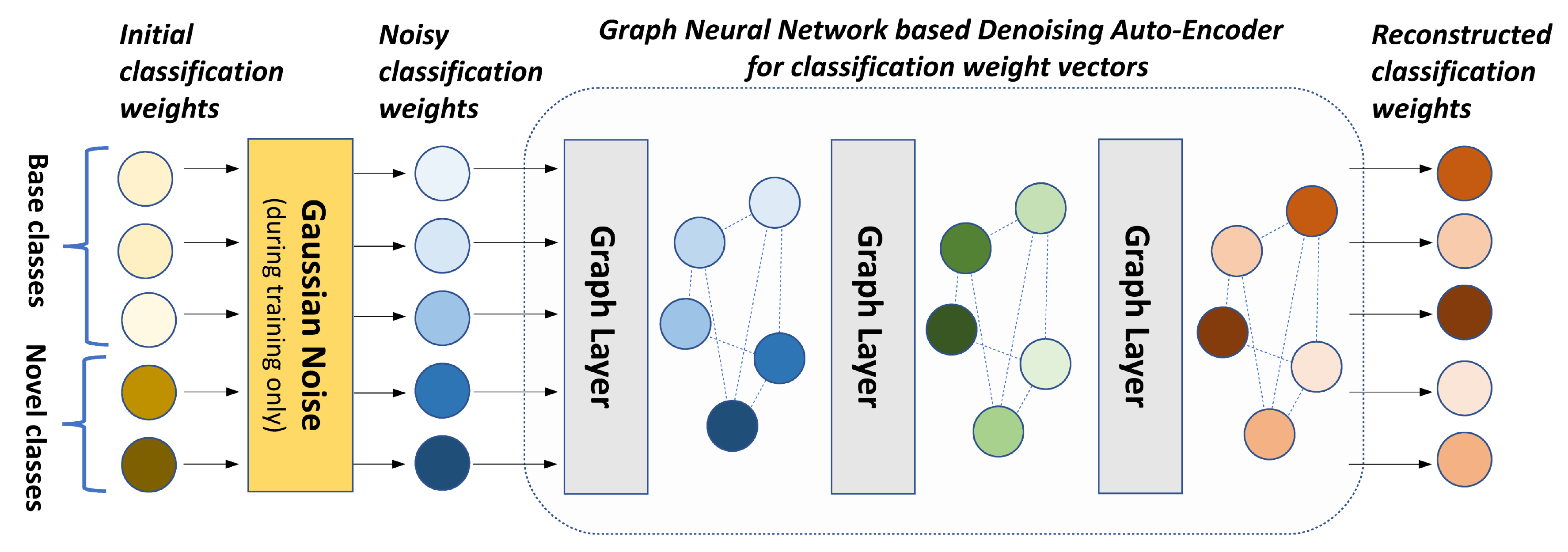}
\end{center}
\caption{
Given a few training data of some novel classes, our meta-model adapts an existing recognition model such that it can classify in a uniﬁed way both novel and base classes by generating classiﬁcation weights for both types of classes. We learn to perform this task by employing a Denoising Autoencoders (DAE) for classiﬁcation weight vectors. Specifically, given some initial estimate of classiﬁcation weights injected with additive Gaussian noise, the DAE is trained to reconstruct target-discriminative classification weights, where the injected noise serves the role of regularizing the weight generation meta-model. Furthermore, in order to capture co-dependencies between different classes (in a given task instance of our meta-model), we implement the DAE model by use of a Graph Neural Network (GNN) architecture.}
\label{fig:overview}
\end{figure*}

We define as $C( F(\cdot|{\theta}) | {\bf w} )$ a recognition model, where 
$F(\cdot | \theta)$ is the feature extractor part of the network with parameters ${\theta}$,
and $C(\cdot | {\bf w})$ is the feature classification part with parameters ${\bf w}$.
The parameters ${\bf w}$ of the classifier consists of $N$ classification weight vectors, 
${\bf w} = \{ {\bf w}_{i} \}_{i=1}^{N}$, 
where $N$ is the number of classes that the network can recognize, and ${\bf w}_i \in \mathbb{R}^{d}$ is the $d$-dimensional classification weight vector of the $i$-th class. 
Given an image ${\bf x}$, the feature extractor will output a $d$-dimensional feature ${\bf z} = F({\bf x}|{\theta})$ and then the classifier will compute the classification scores $[s_1, ..., s_N] = [{\bf z}^\intercal {\bf w}_1, ..., {\bf z}^\intercal {\bf w}_N] \coloneqq C({\bf z} | {\bf w})$ of the $N$ classes.
In our work we use cosine similarity-based feature classifiers~\cite{gidaris2018dynamic, qi2018low}\footnote{This practically means that the features ${\bf z}$ and the classification weights ${\bf w}_i \in {\bf w}$ are $L_2$ normalized} that have been shown to exhibit better performance when transferred to a few-shot recognition task and to be more appropriate for performing unified recognition of both base and novel classes.
Therefore, for the above formulation to be valid, we assume that the features ${\bf z}$ of the feature extractor and the classification weights ${\bf w}_i \in {\bf w}$ of the classifier are already $L_2$ normalized.

Following the formulation of Gidaris and Komodakis~\cite{gidaris2018dynamic},
we assume that the recognition network has already been trained to recognize a set of $N_{bs}$ base classes using a training set $D_{tr}^{bs}$.
The classification weight vectors that correspond to those $N_{bs}$ classes are defined as ${\bf w}^{bs} = \{{\bf w}^{bs}_i\}_{i=1}^{N_{bs}}$.
Our goal is to learn a parameter-generating function $g(\cdot|\phi)$ that, given as input the classification weights ${\bf w}^{bs}$ of the base classes,
and a few training data $D_{tr}^{nv} = \bigcup_{i=N_{bs} + 1}^{N_{bs} + N_{nv}} \{ {\bf x}_{k,i} \}_{k=1}^{K}$ of $N_{nv}$ novel classes, it will be able to output a new set of classification weight vectors
\begin{equation}
{\bf w} = \{ {\bf w}_i \}_{i=1}^{N= N_{bs} + N_{nv}} = g(D_{tr}^{nv}, {\bf w}^{bs} | \phi)
\end{equation}
for both the base and the novel classes, 
where ${K}$ is the number of training examples per novel class, ${\bf x}_{k,i}$ is the $k$-th training example of $i$-th novel, $N = N_{bs} + N_{nv}$ is the total number of classes, and $\phi$ are the learnable parameters of the weight generating function.
This new set of classification weights ${\bf w}$ will be used from the classifier $C(\cdot | {\bf w})$ for recognizing from now on in a unified way both the base and the novel classes.

The parameter-generating function consists of a Denoising Autoencoder for classification weight vectors implemented with a Graph Neural Network (see Figure~\ref{fig:overview} for an overview). 
In the remainder of this section we will describe in more detail how exactly we implement this parameter-generating function.
 
\subsection{Denoising Autoencoders for model parameters generation}
 
In our work we perform the task of generating classification weights by employing a Denoising Autoencoder (DAE) for classification weight vectors.
The injected noise on the classification weights that the DAE framework prescribes serves the role of regularizing the weights generation model $g(\cdot |\phi)$ and thus (as we will see in the experimental section) boost its performance. 
Furthermore, the DAE formulation allows to perform the weights generation task as that of (iteratively) refining some initial (but not very accurate) estimates of the weights in a way that moves them to more probable configurations (when conditioned on the available training data). 
Note that the learnable parameters $\phi$ in $g(\cdot|\phi)$ refer to the learnable parameters of the employed DAE model. In the remainder of this section will briefly provide some preliminaries about DAE models and 
then explain how they are being used in our case.

\paragraph{Preliminaries about DAE.}
Denoising autoencoders are neural networks that, given inputs corrupted by noise, are trained to reconstruct ``clean" versions of them.
By being trained on this task they learn the structure of the data to which they are applied.
It has been shown~\cite{alain2014regularized} that a DAE model,
when trained on inputs corrupted with additive Gaussian noise, can estimate the gradient of the energy function of the density $p({\bf w})$ of its inputs ${\bf w}$:
\begin{equation}
\frac{\partial \, \log \, p({\bf w})}{\partial {\bf w}} \approx \frac{1}{\sigma^2} \cdot (r({\bf w}) - {\bf w}) \textrm{ ,}
\end{equation}
where $\sigma^2$ is the amount of Gaussian noise injected during training, 
and $r(\cdot)$ is the autoencoder.
The approximation becomes exact as $\sigma \rightarrow 0$, and the autoencoder is given enough capacity and training examples.
The direction $\left(r({\bf w}) - {\bf w}\right)$ points towards more likely configurations of ${\bf w}$.  Therefore, the DAE learns a vector field pointing towards the manifold where the input data lies. 
Those theoretical results are independent of the parametrization of the autoencoder.

\paragraph{Applying DAE for classification weight generation.}
In our case we are interested in learning a DAE model that, given an initial estimate of the classification weight vectors ${\bf w}$, would provide a vector field that points towards more probable configurations of ${\bf w}$ conditioned on the training data $D_{tr} = \{D_{tr}^{nv}, D_{tr}^{bs}\}$.
Therefore, we are interested in a DAE model that learns to estimate:
\begin{equation}
\frac{\partial \, \log \, p({\bf w} | D_{tr})}{\partial {\bf w}} \approx \frac{1}{\sigma^2} \cdot (r({\bf w}) - {\bf w}) \textrm{ ,}
\end{equation}
where $p({\bf w} | D_{tr})$ is the conditional distribution of ${\bf w}$ given $D_{tr}$,
    and $r({\bf w})$ is a DAE for the classification weights.
So, after having trained our DAE model for the classification weights, $r({\bf w})$,
we can perform gradient ascent in $\log \, p({\bf w} | D_{tr})$ in order to (iteratively) reach a mode of the estimated conditional distribution $p({\bf w} | D_{tr})$:
\begin{equation} \label{eq:update_rule}
{\bf w} \leftarrow {\bf w} + \epsilon \cdot \frac{\partial \, \log \, p({\bf w} | D_{tr})}{\partial {\bf w}} = {\bf w} + \epsilon \cdot (r({\bf w}) - {\bf w}) \textrm{ ,} 
\end{equation}
where $\epsilon$ is the gradient ascent step size. 

\begin{figure*}[t]
\renewcommand{\figurename}{Figure}
\renewcommand{\captionlabelfont}{\bf}
\renewcommand{\captionfont}{\small} 
\begin{center}
\includegraphics[width=0.90\textwidth]{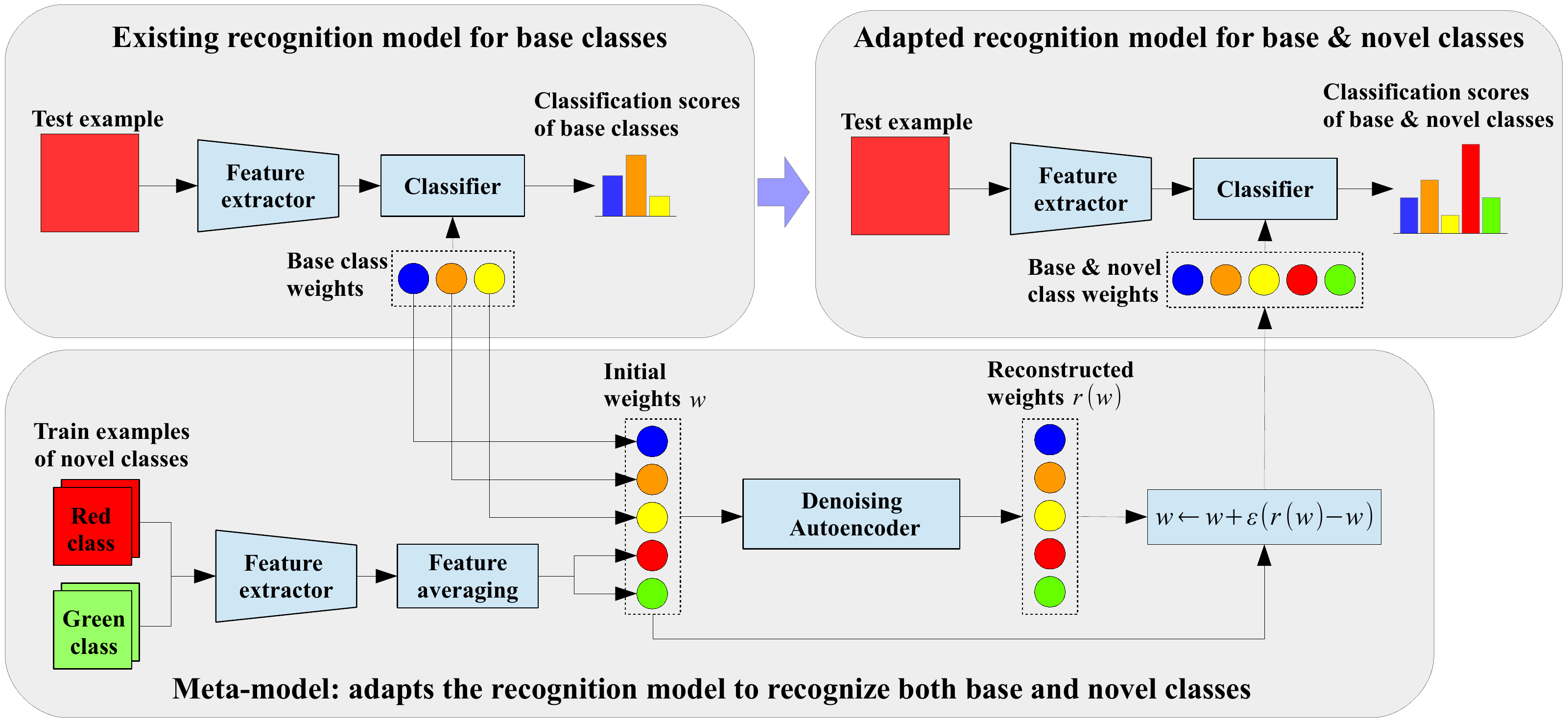}
\end{center}
\caption{
Overview of how our meta-model (bottom) is applied at test time in order to properly adapt an existing recognition model (top-left) into a new model (top-right) that can classify in a unified way both novel and base classes (where only a few training data are provided at test time for the novel classes).}
\label{fig:adaption_overview}
\end{figure*}

The above iterative inference mechanism of the classification weights requires to have available an initial estimate of them. 
This initial estimate is produced using the training data $D_{tr}^{nv}$ of the novel classes and the existing classification weights ${\bf w}^{bs} = \{ {\bf w}^{bs}_i \}_{i=1}^{N_{bs}}$ of the base classes.
Specifically, for the base classes we build this initial estimate by using the classification weights already available in ${\bf w}^{bs}$ and for the novel classes by averaging for each of them the feature vectors of their few training examples:
\begin{equation} \label{eq:initial_w}
{\bf w}_i = 
\begin{cases}
    {\bf w}_i^{bs}, & \text{if $i$ is a base class} \\
    \frac{1}{K} \sum_{k=1}^{K} F({\bf x}_{k,i}|{\theta}),    & \text{otherwise}
\end{cases} ~,
\end{equation}
where ${\bf x}_{k,i}$ is the $k$-th training example of the novel class $i$.

To summarize, our weight generation function $g(D_{tr}^{nv}, {\bf w}^{bs} | \phi)$ is implemented by first producing an initial estimate of the new classification weights by applying equation (\ref{eq:initial_w}) (for which it uses $D_{tr}^{nv}$, and ${\bf w}^{bs}$ ) and then refining those initial estimates by applying the weight update rule of equation (\ref{eq:update_rule}) using the classification weights DAE model $r({\bf w})$ (see an overview of this procedure in Figure~\ref{fig:adaption_overview}).

\paragraph{Episodic training of classification weights DAE model.}
For training, the DAE framework prescribes to apply Gaussian noise on some target weights and then train the DAE model $r(\cdot)$ to reconstruct them.
However, in our case, it is more effective to follow a training procedure that more closely mimics how the DAE model would be used during test time. 
Therefore, we propose to train the DAE model using a learning procedure that is based on training episodes~\cite{vinyals2016matching}.
More specifically, during training we form training episodes by sampling $\tilde{N}_{nv}$ ``fake" novel classes from the available $N_{bs}$ classes in the training data $D_{tr}^{bs}$ and use the remaining $\tilde{N}_{bs} = N_{bs} - \tilde{N}_{nv}$ classes as base ones.
We call the sampled novel classes ``fake" because they actually belong to the set of base classes but during this training episode are treated as being novel.
So, for each ``fake" novel class we sample $K$ training examples forming a training set for the ``fake" novel classes of this training episode.
We also sample $M$ training examples both from the ``fake" novel and the remaining base classes forming the validation set $D_{val} = \{ ({\bf x}_m, y_m) \}_{m=1}^{M}$ of this training episode, where $({\bf x}_m, y_m)$ is the image ${\bf x}_m$ and the label $y_m$ of the $m$-th validation example.
Then, we produce an initial estimate ${\bf \tilde{w}}$ of the classification weights of the sampled classes (both ``fake" novel and base) using the mechanics of equation (\ref{eq:initial_w}), and we inject to it Gaussian additive noise ${\bf \tilde{w}} = \{ {\bf \tilde{w}}_i + \varepsilon \}_{i=1}^{\tilde{N} = \tilde{N}_{bs} +  \tilde{N}_{nv}}$, where $\varepsilon \sim N(0, \sigma)$.
We give ${\bf \tilde{w}}$ as input to the DAE model in order to output the reconstructed weights ${\bf \hat{w}} = \{ {\bf \hat{w}}_i \}_{i=1}^{\tilde{N}}$. 
Having computed ${\bf \hat{w}}$ we apply to them a squared reconstruction loss of the target weights ${\bf w^*} = \{ {\bf w^*}_i \}_{i=1}^{\tilde{N}}$ and a classification loss of the $M$ validation examples of this training episode:
\begin{equation} \label{eq:loss}
\frac{1}{N} \sum_{i=1}^{\tilde{N}} \| {\bf \hat{w}}_i - {\bf w^*}_i \|^2 + \frac{1}{M} \sum_{m=1}^{M} loss({\bf x}_m, y_m | \hat{{\bf w}})~,
\end{equation}
where $loss({\bf x}_m, y_m | {\bf\hat{w}}) = - {\bf z}_m^\intercal {\bf \hat{w}}_{y_m} + \log(\sum_{i=1}^{\tilde{N}} e^{{\bf z}_m^\intercal {\bf \hat{w}}_{i}})$ is the cross entropy loss of the $m$-th validation example and ${\bf z}_m = F({\bf x}_m|{\theta})$ is the feature vector of the $m$-th example.
Note that the target weights ${\bf w^*}$ are the corresponding base class weight vectors already learned by the recognition model.

\subsection{Graph Neural Network based Denoising Autoencoder} \label{sec:gnn}

Here we describe how we implement our DAE model.
Reconstructing the classification weights of the novel classes, for which training data are scarce, is an ill-defined problem. 
One way to boost DAE's reconstruction performance is by allowing it to take into account the inter-class relationships when reconstructing the classification weights of a set of classes. 
Given the unordered nature in a set of classes, we chose to implement the DAE model by use of a Graph Neural Network (GNN).
In the remainder of this subsection we describe how we employed a GNN for the reconstruction task and what type of GNN architecture we used.

\begin{figure}[t]
\renewcommand{\figurename}{Figure}
\renewcommand{\captionlabelfont}{\bf}
\renewcommand{\captionfont}{\small} 
\begin{center}
\begin{subfigure}[b]{0.48\textwidth}
        \includegraphics[width=1.00\textwidth]{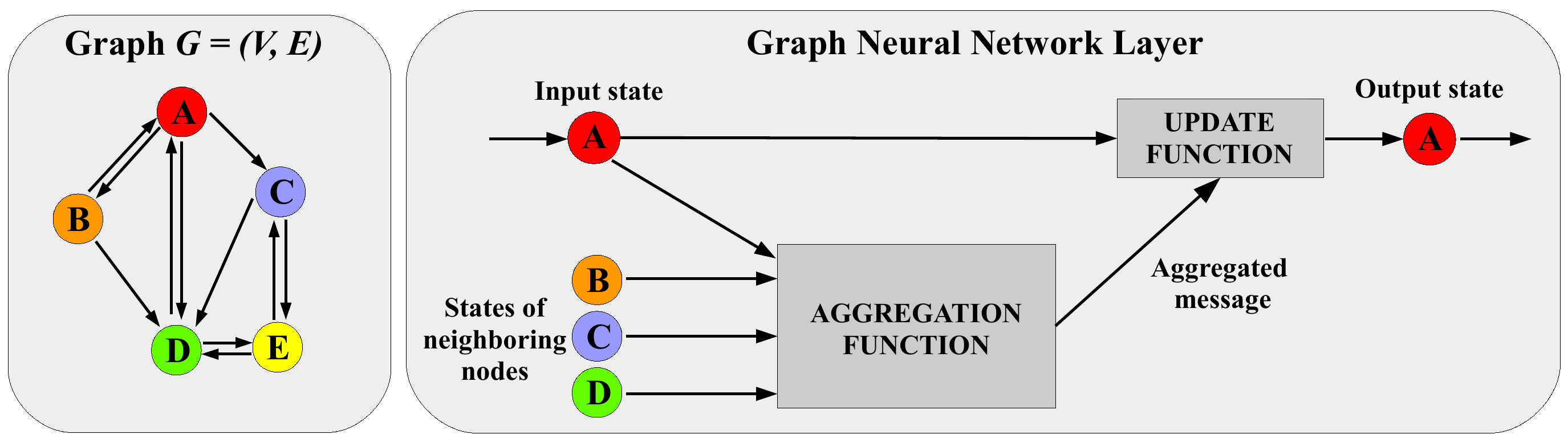}
        \caption{The general architecture of a GNN layer.}
        \label{fig:generic_gnn_layer}
\end{subfigure}
\begin{subfigure}[b]{0.48\textwidth}
        \includegraphics[width=1.00\textwidth]{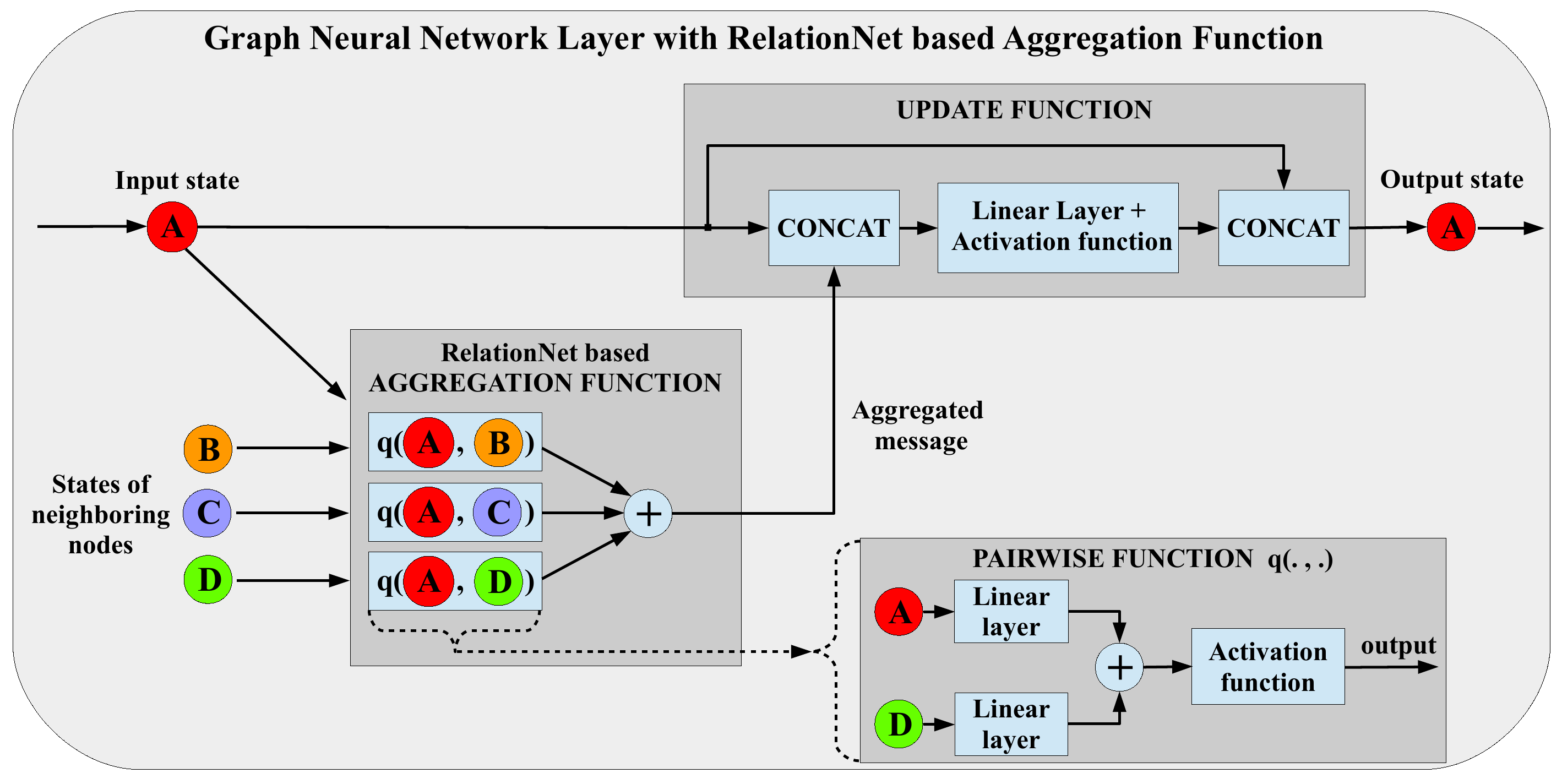}
        \caption{The architecture of the hidden GNN layers in our work.}
        \label{fig:relation_gnn_layer}
\end{subfigure}
\end{center}
\caption{
\textbf{(a)} A GNN layer typically consists of two functions, an \emph{aggregation} function that, for a node of interest (e.g., node A in the figure), aggregates information from its neighboring nodes, and an \emph{update} function that updates the state (i.e., the features) of that node by taking into account both the state of that node and the aggregated information from its neighborhood.
\textbf{(b)} The GNN layer architecture that we use in our work implements the aggregation function as a small Relation-Net network~\cite{santoro2017simple}. The two linear layers in the pairwise function $q(\cdot,\cdot)$ are the same (i.e., share parameters).}
\label{fig:gnn_layer}
\end{figure}

GNNs are multi-layer networks that operate on graphs $G = (V, E)$ by structuring their computations according to the graph connectivity.
i.e., at each GNN layer the feature responses of a node are computed based on the neighboring nodes defined by the adjacency graph (see Figure~\ref{fig:generic_gnn_layer} for an illustration).
In our case, we associate the set of classes $Y = \{i\}_{i=1}^{N}$ 
(for which we want to reconstruct their classification weights) 
with a graph $G = (V, E)$, where each node $v_i \in V$ corresponds to the class $i$ in $Y$ (either base or novel).
To define the set of edges $(i,j) \in E$ of the graph,
we connect each class $i$ with its $J$\footnote{In our experiments we use $J=10$ classes as neighbors.} closest classes in terms of the cosine-similarity of the initial estimates of their classification weight vectors (before the injection of the Gaussian noise).
The edge strength $a_{ij} \in [0, 1]$ of each edge $(i,j) \in E$ is computed by applying the softmax operation over the cosine similarities scores of its neighbors $\mathcal{N}(i) = \{ j, \forall (i,j) \in E \}$, thus forcing $\sum_{j \in \mathcal{N}(i)} a_{ij}=1$\footnote{For this softmax operation we used an inverse temperature value of $5$.}. 
We define as ${\bf h}^{(l)} = \{{\bf h}_i^{(l)}\}_{i=1}^{N}$ the set of feature vectors that represent the $N$ graph nodes (i.e., the $N$ classes) at the $l$-th level of the GNN. 
In this case, the input set ${\bf h}^{(0)}$ to the GNN is the set of classification weight vectors ${\bf w} = \{ {\bf w}_{i} \}_{i=1}^{N} = {\bf h}^{(0)}$ that the GNN model will refine.
Each GNN layer, receives as input the set ${\bf h}^{(l)}$ and outputs a new set ${\bf h}^{(l+1)}$ as:
\begin{equation} \label{eq:gnn_aggregate}
{\bf h}_{\mathcal{N}(i)}^{(l)} = \texttt{AGGREGATE} \left(\{ {\bf h}_{j}^{(l)}, \forall j \in \mathcal{N}(i)\} \right)~,
\end{equation}
\begin{equation} \label{eq:gnn_unify}
{\bf h}_i^{(l+1)} = \texttt{UPDATE} \left({\bf h}_i^{(l)}, {\bf h}_{\mathcal{N}(i)}^{(l)} \right)~,
\end{equation}
where $\texttt{AGGREGATE}(\cdot)$ is a parametric function that for each node $i$ aggregates information from its node neighbors $\mathcal{N}(i)$ to generate the message feature ${\bf h}_{\mathcal{N}(i)}^{(l)}$, 
and $\texttt{UPDATE}(\cdot, \cdot)$ is a parametric function that for each node $i$ will get as input the features ${\bf h}_i^{(l)}$, ${\bf h}_{\mathcal{N}(i)}^{(l)}$ and will compute the new feature vector ${\bf h}_i^{(l+1)}$ of that node.

\paragraph{Relation-Net based aggregation function.}
Generally,
the aggregation function is implemented as linear combination of message vectors received from the node neighbors:
\begin{equation} \label{eq:gnn_relationet_aggregation}
{\bf h}_{\mathcal{N}(i)}^{(l)} = \sum_{j \in \mathcal{N}(i)} a_{ij} \cdot q^{(l)}\left({\bf h}_i^{(l)}, {\bf h}_j^{(l)}\right)~, 
\end{equation}
where $q^{(l)}\left({\bf h}_i^{(l)}, {\bf h}_j^{(l)}\right)$ is a function that computes the message vector that node $i$ receives from its neighbor $j$.
Inspired by relation networks~\cite{santoro2017simple}, we implement $q^{(l)}\left({\bf h}_i^{(l)}, {\bf h}_j^{(l)}\right)$ as a non-linear parametric function of the feature vectors of both the sender and the receiver nodes.
Specifically, 
given the two input vectors, $q^{(l)}$ forwards each of them through the same fully connected linear layer, adds their outputs, and then applies BatchNorm + Dropout + LeakyReLU units (see Figure~\ref{fig:relation_gnn_layer}). 
Note that in this implementation the message between two nodes, is independent of the direction of the message, i.e., $q^{(l)}\left({\bf h}_i^{(l)}, {\bf h}_j^{(l)}\right) = q^{(l)}\left({\bf h}_j^{(l)}, {\bf h}_i^{(l)}\right)$.

\paragraph{Update function.}
The update function of the \emph{hidden GNN layers} is implemented as:
\begin{equation} \label{eq:gnn_hidden_update1}
{\bf h}_i^{(l+1)} = \left[{\bf h}_i^{(l)};~u^{(l)} \left(\left[{\bf h}_i^{(l)};~{\bf h}_{\mathcal{N}(i)}^{(l)} \right]\right) \right]~,
\end{equation}
where $\left[\pmb{\alpha}; \pmb{\beta} \right]$ is the concatenation of vectors $\pmb{\alpha}$ and $\pmb{\beta}$, and
$u^{(l)}(\cdot)$ is a non-linear parametric function that, 
given as input a vector, forwards it through a fully connected linear layer followed by BatchNorm + Dropout + LeakyReLU + $L_2$-normalization units (see Figure~\ref{fig:relation_gnn_layer}).
For the \emph{last prediction GNN layer}, the update function is implemented as:
\begin{equation} \label{eq:gnn_final_update}
{\bf \delta w}_i,~{\bf o}_i =u^{(L-1)} \left(\left[{\bf h}_i^{(L-1)};~ {\bf h}_{\mathcal{N}(i)}^{(L-1)} \right]\right)~,
\end{equation}
where $u^{(L-1)}(\cdot)$ is a non-linear parametric function that, given an input vector, outputs the two $d$-dimensional vectors ${\bf \delta w}_i$ and ${\bf o}_i$.
$u^{(L-1)}(\cdot)$ is implemented as a fully connected linear layer followed by a $L_2$-normalization unit for the ${\bf \delta w}_i$ output, and a Sigmoid unit for the ${\bf o}_i$ output.
The final output of the GNN is computed with the following operation:
\begin{equation} \label{eq:gnn_final_output2}
{\bf \hat{w}}_i = {\bf w}_i + {\bf o}_i \odot {\bf \delta w}_i.
\end{equation}
As can be seen, we opted for residual-like predictions of the new classification weights ${\bf \hat{w}}_i$, 
since this type of operations are more appropriate for the type of refining/denoising that must be performed by our DAE model.
Our particular implementation uses the gating vectors ${\bf o}_i$ to control the amount of contribution of the residuals ${\bf \delta w}_i$ to the input weights ${\bf w}_i$.

We named this GNN based DAE model for weights reconstruction \emph{wDAE-GNN} model.
Alternatively, we also explored a simpler DAE model that is implemented to reconstruct each classification weight vector (in a given task instance of our meta-model) independently with a MLP network (\emph{wDAE-MLP} model).
More specifically, the \emph{wDAE-MLP} model is implemented with layers similar to those of the GNN that \emph{only include the update function part and not the aggregation function part}.
So, it only includes fully connected layers and the skip connections (i.e., 2nd and 3rd boxes in update function part of Figure~\ref{fig:relation_gnn_layer}).

\section{Experimental Evaluation} \label{sec:experimentalresults}
\begin{table*}[t!]
\centering
\renewcommand{\figurename}{Table}
\renewcommand{\captionlabelfont}{\bf}
\renewcommand{\captionfont}{\small} 
\renewcommand{\arraystretch}{1.2}
\renewcommand{\tabcolsep}{1.2mm}
\resizebox{0.7\linewidth}{!}{
\begin{tabular}{l |@{\hspace{4mm}}ccccc@{\hspace{4mm}} |@{\hspace{4mm}}ccccc@{\hspace{4mm}}}
& \multicolumn{5}{@{\hspace{4mm}}c@{\hspace{4mm}}}{Novel classes} & \multicolumn{5}{@{\hspace{4mm}}c@{\hspace{4mm}}}{All classes}\\
Approach &$K$=1 & 2 & 5 & 10 & 20&$K$=1 & 2 & 5 & 10 & 20\\
\midrule
\multicolumn{11}{l}{\textbf{\emph{Prior work}}}\\
\;Prototypical-Nets (from \cite{wang2018low})                & 39.3 & 54.4 & 66.3 & 71.2 & 73.9 & 49.5 & 61.0 & 69.7 & 72.9 & 74.6\\
\;Matching Networks (from \cite{wang2018low})                & 43.6 & 54.0 & 66.0 & 72.5 & 76.9 & 54.4 & 61.0 & 69.0 & 73.7 & 76.5\\
\;Logistic regression \cite{hariharan2016low}                & 38.4 & 51.1 & 64.8 & 71.6  & 76.6 & 40.8 & 49.9 & 64.2 & 71.9 & 76.9\\
\;Logistic regression w/ H \cite{hariharan2016low}           & 40.7 & 50.8 & 62.0 & 69.3 & 76.5 & 52.2 & 59.4 & 67.6 & 72.8 & 76.9\\
\;SGM w/ H \cite{hariharan2016low}                           & - & - & - & - & - & 54.3 & 62.1 & 71.3 & 75.8 & 78.1\\
\;Batch SGM \cite{hariharan2016low}                          & - & - & - & - & - & 49.3 & 60.5 & 71.4 & 75.8 & 78.5\\
\;Prototype Matching Nets w/ H \cite{wang2018low}            & 45.8 & 57.8 & 69.0 & 74.3 & 77.4 & 57.6 & 64.7 & 71.9 & 75.2 & 77.5\\
\;LwoF \cite{gidaris2018dynamic}  & 46.2 & 57.5 & 69.2 & 74.8 & \textbf{78.1} & 58.2 & 65.2 & 72.7 & \textbf{76.5} & \textbf{78.7}\\
\midrule
\multicolumn{11}{l}{\textbf{\emph{Ours}}}\\
\;wDAE-GNN                 & \textbf{48.0} & \textbf{59.7} & \textbf{70.3} & \textbf{75.0} & 77.8 & \textbf{59.1} & \textbf{66.3} & \textbf{73.2} & 76.1 & 77.5\\
\;wDAE-MLP                 & 47.6 & 59.2 & 70.0 & 74.8 & 77.7 & 59.0 & 66.1 & 72.9 & 75.8 & 77.4\\
\midrule
\multicolumn{11}{l}{\textbf{\emph{Ablation study on wDAE-GNN}}}\\
\;Initial estimates               & 45.4 & 56.9 & 68.9 & 74.5 & 77.7 & 57.0 & 64.3 & 72.3 & 75.6 & 77.3\\
\;wDAE-GNN - No Noise               & 47.6 & 59.0 & 70.0 & 74.9 & 77.8 & 60.0 & 66.0 & 72.9 & 75.8 & 77.4\\
\;wDAE-GNN - Noisy Targets as Input & 47.8 & 59.4 & 70.1 & 74.8 & 77.7 & 58.7 & 66.0 & 73.1 & 76.0 & 77.5\\
\;wDAE-GNN - No Cls. Loss           & 47.7 & 59.1 & 69.8 & 74.6 & 77.6 & 58.4 & 65.5 & 72.7 & 75.8 & 77.5\\
\;wDAE-GNN - No Rec. Loss           & 47.8 & 59.4 & 70.1 & 75.0 & 77.8 & 58.7 & 66.0 & 73.1 & 76.1 & 77.6\\
\bottomrule
\end{tabular}
}
\caption{
Top-5 accuracies on the novel and on all classes for the ImageNet-FS benchmark~\cite{hariharan2016low}.
To report results we use 100 test episodes. For all our models the $95\%$ confidence intervals on the $K=1$, $2$, $5$, $10$, and $20$ settings are (around) $\pm 0.21$, $\pm 0.15$, $\pm 0.08$, $\pm 0.06$, and $\pm 0.05$ respectively for Novel classes and $\pm 0.13$, $\pm 0.10$, $\pm 0.05$, $\pm 0.04$, and $\pm 0.03$ for All classes.}
\label{tab:lowshotTop5}
\end{table*}
\begin{table}[t]
\centering
\renewcommand{\figurename}{Table}
\renewcommand{\captionlabelfont}{\bf}
\renewcommand{\captionfont}{\small} 
\resizebox{1.0\linewidth}{!}{
{\setlength{\extrarowheight}{2pt}\scriptsize
{
\begin{tabular}{l <{\hspace{-0.3em}} | >{\hspace{-0.5em}} l | >{\hspace{-0.5em}} l | >{\hspace{-0.5em}} l}
\toprule
\multicolumn{1}{l|}{Models} & Backbone & \multicolumn{1}{c|}{1-shot} & \multicolumn{1}{c}{5-shot}\\
\midrule
\multicolumn{4}{l}{\textbf{\emph{Prior work}}}\\
\;MAML~\cite{finn2017model}                           & Conv-4-64  & 48.70 $\pm$ 1.84\% & 63.10 $\pm$ 0.92\%\\
\;Prototypical Nets~\cite{snell2017prototypical}      & Conv-4-64  & 49.42 $\pm$ 0.78\% & 68.20 $\pm$ 0.66\%\\
\;LwoF~\cite{gidaris2018dynamic}                      & Conv-4-64  & 56.20 $\pm$ 0.86\% & 72.81 $\pm$ 0.62\%\\
\;RelationNet~\cite{yang2018learning}                 & Conv-4-64  & 50.40 $\pm$ 0.80\% & 65.30 $\pm$ 0.70\%\\
\;GNN~\cite{garcia2017few}                            & Conv-4-64  & 50.30\%          & 66.40\%\\
\;R2-D2~\cite{bertinetto2018meta}                     & Conv-4-64  & 48.70 $\pm$ 0.60\% & 65.50 $\pm$ 0.60\%\\
\;R2-D2~\cite{bertinetto2018meta}                     & Conv-4-512 & 51.20 $\pm$ 0.60\% & 68.20 $\pm$ 0.60\%\\
\;TADAM~\cite{oreshkin2018tadam}                      & ResNet-12  & 58.50 $\pm$ 0.30\% & 76.70 $\pm$ 0.30\%\\
\;Munkhdalai~\etal\cite{munkhdalai2017meta}           & ResNet-12  & 57.10 $\pm$ 0.70\% & 70.04 $\pm$ 0.63\%\\
\;SNAIL~\cite{santoro2017simple}                      & ResNet-12  & 55.71 $\pm$ 0.99\% & 68.88 $\pm$ 0.92\%\\
\;Qiao~\etal\cite{qiao2017few}$^\dagger$              & WRN-28-10 & 59.60 $\pm$ 0.41\% & 73.74 $\pm$ 0.19\%\\
\;LEO~\cite{rusu2018meta}$^\dagger$                   & WRN-28-10 & 61.76 $\pm$ 0.08\% & 77.59 $\pm$ 0.12\%\\
\;LwoF~\cite{gidaris2018dynamic} (our implementation) & WRN-28-10 & 60.06 $\pm$ 0.14\% & 76.39 $\pm$ 0.11\% \\
\midrule
\multicolumn{4}{l}{\textbf{\emph{Ours}}}\\
\;wDAE-GNN                                            & WRN-28-10 & 61.07 $\pm$ 0.15\% & 76.75 $\pm$ 0.11\% \\
\;wDAE-MLP                                            & WRN-28-10 & 60.61 $\pm$ 0.15\% & 76.56 $\pm$ 0.11\% \\
\;wDAE-GNN$^\dagger$                                  & WRN-28-10 & \textbf{62.96} $\pm$ 0.15\% & \textbf{78.85} $\pm$ 0.10\% \\
\;wDAE-MLP$^\dagger$                                  & WRN-28-10 & 62.67 $\pm$ 0.15\% & 78.70 $\pm$ 0.10\% \\
\midrule
\multicolumn{4}{l}{\textbf{\emph{Ablation study on wDAE-GNN}}}\\
\;Initial estimate                                    & WRN-28-10 & 59.68 $\pm$ 0.14\% & 76.48 $\pm$ 0.11\% \\
\;wDAE-GNN - No Noise                                 & WRN-28-10 & 60.29 $\pm$ 0.14\% & 76.49 $\pm$ 0.11\% \\
\;wDAE-GNN - Noisy Targets as Input                   & WRN-28-10 & 60.92 $\pm$ 0.15\% & 76.69 $\pm$ 0.11\% \\
\;wDAE-GNN - No Cls. Loss                             & WRN-28-10 & 60.96 $\pm$ 0.15\% & 76.75 $\pm$ 0.11\% \\
\;wDAE-GNN - No Rec. Loss                             & WRN-28-10 & 60.76 $\pm$ 0.15\% & 76.64 $\pm$ 0.11\% \\
\midrule
\multicolumn{4}{l}{\textbf{\emph{Ablation study on wDAE-MLP}}}\\
\;wDAE-MLP - No Noise                                 & WRN-28-10 & 60.16 $\pm$ 0.15\% & 76.50 $\pm$ 0.11\% \\
\;wDAE-MLP - Noisy Targets as Input                   & WRN-28-10 & 60.43 $\pm$ 0.15\% & 76.49 $\pm$ 0.11\% \\
\;wDAE-MLP - No Cls. Loss                             & WRN-28-10 & 60.55 $\pm$ 0.15\% & 76.62 $\pm$ 0.11\% \\
\;wDAE-MLP - No Rec. Loss                             & WRN-28-10 & 60.45 $\pm$ 0.15\% & 76.50 $\pm$ 0.11\% \\
\bottomrule
\end{tabular}}}}
\caption{
Top-1 accuracies on the novel classes of MiniImageNet test set with $95\%$ confidence intervals.
$^\dagger$: using also the validation classes for training.}
\label{tab:MiniImagenetResults}
\end{table}

In this section we first compare our method against prior work in~\S\ref{sec:comparison} and then in~\S\ref{sec:analysis} we perform a detailed experimental analysis of it. 

\paragraph{Datasets and Evaluation Metrics.}
We evaluate our approach on three datasets, ImageNet-FS~\cite{hariharan2016low, wang2018low}, MiniImageNet~\cite{vinyals2016matching}, and \emph{tiered}-MiniImageNet~\cite{ren2018meta}.
\textbf{ImageNet-FS} is a few-shot benchmark created by splitting the ImageNet classses~\cite{russakovsky2015imagenet} into $389$ base classes and $611$ novel classes; 
$193$ of the base classes and $300$ of the novel classes are used for validation and the remaining $196$ base classes and $311$ novel classes are used for testing.
In this benchmark the models are evaluated based on (1) the recognition performance of the $311$ test novel classes (i.e., $311$-way classification task), and 
(2) recognition of all the $507$ classes (i.e., both the $196$ test base classes and the $311$ novel classes; for more details we refer to~\cite{hariharan2016low, wang2018low}).
We report result for $K=1$, $2$, $5$, $10$, or $20$ training examples per novel class. 
For each of those $K$-shot settings we sample $100$ test episodes (where each episode consists of sampling $K$ training examples per novel class and then evaluating on the validation set of ImageNet) and compute the mean accuracies over all episodes.
\textbf{MiniImageNet} consists of $100$ classes randomly picked from the ImageNet with $600$ images per class. 
Those $100$ classes are divided into $64$ base classes, $16$ validation novel classes, and $20$ test novel classes. The images in MiniImageNet have size $84 \times 84$ pixels.
\textbf{\emph{tiered}-MiniImageNet} consists of ImageNet $608$ classes divided into $351$ base classes, $97$ validation novel classes, and $160$ novel test classes.
In total there are $779,165$ images again with size $84 \times 84$.
In MiniImageNet and \emph{tiered}-MiniImageNet, 
the models are evaluated on several $5$-way classification tasks (i.e., test episodes) created by first randomly sampling $5$ novel classes from the available test novel classes, and then $K=1$, or $5$ training examples and $M=15$ test examples per novel class.
To report results we use $20000$ such test episodes and compute the mean accuracies over all episodes.
Note that when learning the novel classes we also feed the base classes to our \emph{wDAE-GNN} models in order to take into account the class dependencies between the novel and base classes.

\subsection{Implementation details}

\paragraph{Feature extractor architectures.}
For the ImageNet-FS experiments we use a ResNet-10~\cite{he2016deep} architecture that given images of size $224 \times 224$ outputs $512$-dimensional feature vectors.
For the MiniImageNet and \emph{tiered}-MiniImageNet experiments we used a 2-layer Wide Residual Network~\cite{zagoruyko2016wide} (WRN-28-10) that receives images of size $80 \times 80$ (resized form $84 \times 84$) and outputs $640$-dimensional feature vectors.

\paragraph{wDAE-GNN and wDAE-MLP architectures.}
In all our experiments we use a wDAE-GNN architecture with two GNN layers.
In ImageNet-FS (MiniImageNet) the $q^{(l)}(\cdot,\cdot)$ parametric function of all GNN layers and the $u^{(l)}(\cdot)$ parametric function of the hidden GNN layer output features of $1024$ ($320$) channels. All dropout units have $0.7$ ($0.95$) dropout ratio, and the $\sigma$ of the Gaussian noise in DAE is $0.08$ ($0.1$). A similar architecture is used in wDAE-MLP but without the aggregation function parts.

For training we use SGD optimizer with momentum $0.9$ and weight decay $5e-4$.
We train our models only on the $1$-shot setting and then use them for all the $K$-shot settings. During test time we apply only 1 refinement step of the initial estimates of the classification weights (i.e., only 1 application of the update rule (\ref{eq:update_rule})).
In ImageNet-FS the step size $\varepsilon$ of the update rule (\ref{eq:update_rule}) is set to $1.0$, $1.0$, $0.6$, $0.4$, and $0.2$ for the $K=1$, $2$, $5$, $10$, and $20$ shot settings respectively.
In MiniImageNet $\varepsilon$ is set to $1.0$ and $0.5$ for the $K=1$ and $K=5$ settings respectively.
All hyper-parameters were cross-validated in the validation splits of the datasets.

We provide the implementation code at \url{https://github.com/gidariss/wDAE_GNN_FewShot}

\subsection{Comparison with prior work} \label{sec:comparison}

Here we compare our \emph{wDAE-GNN} and \emph{wDAE-MLP} models against prior work on the ImageNet-FS, MiniImageNet, and \emph{tiered}-MiniImageNet datasets.
More specifically, on ImageNet-FS (see Table~\ref{tab:lowshotTop5}) the proposed models achieve in most cases superior performance than prior methods - especially on the challenging and interesting scenarios of having less than $5$ training examples per novel class (i.e., $K \le 5$).
For example, the \emph{wDAE-GNN} model improves the $1$-shot accuracy for novel classes of the previous state-of-the-art~\cite{gidaris2018dynamic} by around $1.8$ accuracy points.
On MiniImageNet and \emph{tiered}-MiniImageNet (see Tables~\ref{tab:MiniImagenetResults} and~\ref{tab:tieredMiniImagenet} respectively) the proposed models surpass the previous methods on all settings and achieve new state-of-the-art results.
Also, for MiniImageNet we provide in Table~\ref{tab:MiniImagenetResultsAll} the classification accuracies of both the novel and base classes and we compare with the LwoF~\cite{gidaris2018dynamic} prior work.
Again, we observe that our models surpass the prior work.

\subsection{Analysis of our method} \label{sec:analysis}

\paragraph{Ablation study of DAE framework.}
Here we perform ablation study of various aspects of our DAE framework on the ImageNet and the MiniImageNet datasets (see corresponding results on Tables~\ref{tab:lowshotTop5} and~\ref{tab:MiniImagenetResults}).
Specifically, we examine the cases of 
\textbf{(1)} training the reconstruction models without noise (entries with suffix \emph{No Noise}),
\textbf{(2)} during training providing as input to the model noisy versions of the target classification weights that has to reconstruct (entries with suffix \emph{Noisy Targets as Input}),
\textbf{(3)} training the models without classification loss on the validation examples (i.e., using only the first term of the loss (\ref{eq:loss}); entries with suffix \emph{No Cls. Loss}), and
\textbf{(4)} training the models with only the classification loss on the validation examples and without the reconstruction loss (i.e., using only the second term of the loss (\ref{eq:loss}); entries with suffix \emph{No Rec. Loss}).
\textbf{(5)} We also provide the recognition performance of the initial estimates of the classification weight vectors without being refined by our DAE models (entries \emph{Initial Estimates}).
We observe that each of those ablations to our DAE models lead to worse few-shot recognition performance.
Among them, the models trained without noise on theirs inputs achieves the worst performance, which demonstrates the necessity of the DAE formulation.

\paragraph{Impact of GNN architecture.}
By comparing the classification performance of the \emph{wDAE-GNN} models with the \emph{wDAE-MLP} models in Tables~\ref{tab:lowshotTop5} and~\ref{tab:MiniImagenetResults}, we observe that indeed, taking into account the inter-class dependencies with the proposed GNN architecture is beneficial to the few-shot recognition performance.
Specifically, the GNN architecture offers a small (e.g., around 0.40 percentage points in the 1-shot case) but consistent improvement that according to the confidences intervals of Tables~\ref{tab:lowshotTop5} and~\ref{tab:MiniImagenetResults} is in almost all cases statistically significant.

\section{Conclusion} \label{sec:conclusion}
\begin{table}[t]
\centering
\renewcommand{\figurename}{Table}
\renewcommand{\captionlabelfont}{\bf}
\renewcommand{\captionfont}{\small} 
\resizebox{\linewidth}{!}{
{\setlength{\extrarowheight}{2pt}\scriptsize
{
\begin{tabular}{l <{\hspace{-0.3em}} | >{\hspace{-0.5em}} l | >{\hspace{-0.5em}} l | >{\hspace{-0.5em}} l | >{\hspace{-0.5em}} l }
\toprule
 & \multicolumn{2}{c|}{Novel classes} & \multicolumn{2}{c}{All classes}\\
\multicolumn{1}{l|}{Models} & \multicolumn{1}{c|}{1-shot} & \multicolumn{1}{c|}{5-shot} & \multicolumn{1}{c|}{1-shot} & \multicolumn{1}{c}{5-shot}\\
\midrule
\;LwoF~\cite{gidaris2018dynamic} & 60.03 $\pm$ 0.14\% & 76.35 $\pm$ 0.11\% & 55.70 $\pm$ 0.08\% & 66.27$\pm$ 0.07\%\\
\;wDAE-GNN                       & \textbf{61.07} $\pm$ 0.15\% & \textbf{76.75} $\pm$ 0.11\% & \textbf{56.55} $\pm$ 0.08\%  & 67.00 $\pm$ 0.07\%\\
\;wDAE-MLP                       & 60.61 $\pm$ 0.14\% & 76.56 $\pm$ 0.11\% & 56.07 $\pm$ 0.08\% & \textbf{67.05} $\pm$ 0.07\%\\
\bottomrule
\end{tabular}}}}
\caption{
Top-1 accuracies on the novel and on all classes of MiniImageNet test set with $95\%$ confidence intervals.}
\label{tab:MiniImagenetResultsAll}
\end{table}
\begin{table}[t]
\centering
\renewcommand{\figurename}{Table}
\renewcommand{\captionlabelfont}{\bf}
\renewcommand{\captionfont}{\small} 
\resizebox{\linewidth}{!}{
{\setlength{\extrarowheight}{2pt}\scriptsize
{
\begin{tabular}{l <{\hspace{-0.3em}} | l <{\hspace{-0.3em}} | >{\hspace{-0.5em}} l | >{\hspace{-0.5em}} l}
\toprule
\multicolumn{1}{l|}{Models} & Backbone & \multicolumn{1}{c|}{1-shot} & \multicolumn{1}{c}{5-shot}\\
\midrule
\;MAML~\cite{finn2017model} (from~\cite{liu2018transductive})  & Conv-4-64          &  51.67 $\pm$ 1.81\% &  70.30 $\pm$ 0.08\% \\
\;Prototypical Nets~\cite{snell2017prototypical}  & Conv-4-64                       &  53.31 $\pm$ 0.89\% &  72.69 $\pm$ 0.74 \% \\
\;RelationNet~\cite{yang2018learning} (from~\cite{liu2018transductive}) & Conv-4-64 &  54.48 $\pm$ 0.93\% &  71.32 $\pm$ 0.78\%\\
\;Liu~\etal~\cite{liu2018transductive} & Conv-4-64                                  &  57.41 $\pm$ 0.94\% & 71.55 $\pm$ 0.74\\
\;LEO~\cite{rusu2018meta}   & WRN-28-10                                             &  66.33 $\pm$ 0.05\% &  81.44 $\pm$ 0.09 \%\\
\;LwoF~\cite{gidaris2018dynamic} (our implementation)                               & WRN-28-10 & 67.92 $\pm$ 0.16\% & \textbf{83.10} $\pm$ 0.12\% \\
\midrule
\;wDAE-GNN (Ours)                                                       & WRN-28-10 & \textbf{68.18} $\pm$ 0.16\% & 83.09 $\pm$ 0.12\%\\
\bottomrule
\end{tabular}}}}
\caption{Top-1 accuracies on the novel classes of \emph{tiered}-MiniImageNet test set with $95\%$ confidence intervals.}
\label{tab:tieredMiniImagenet}
\end{table}

We proposed a meta-model for few-shot learning that takes as input a set of novel classes (with few training examples for each of them) and then generates classification weight vectors for them. 
Our model is based on the use of a Denoising Autoencoder (DAE) network.
During training, the injected noise on the classification weights given as input to the DAE network allows the regularization of the learning procedure and helps in boosting the performance of the meta-model.
After training, the DAE model is used for refining an initial set of classification weights with the goal of making them more discriminative with respect to the classification task at hand.
We implemented the above DAE model by use of a Graph Neural Network architecture so as to allow our meta-model to properly learn (and take advantage of) the structure of the \emph{entire} set of classification weights that must be reconstructed on each instance (i.e., episode) of the meta-learning task.
Our detailed experiments on the ImageNet-FS~\cite{hariharan2016low} and MiniImageNet~\cite{vinyals2016matching} datasets reveal 
(1) the significance of our DAE formulation for training meta-models capable to generate classification weights, and 
(2) that the GNN architecture manages to offer a consistent improvement on the few-shot classification accuracy.
Finally, our model surpassed prior methods on all the explored datasets.

\section{Acknowledgements} \label{sec:acknowledgements}
We would like to thank Martin Simonovsky and Shell Xu for fruitful discussions and Thibault Groueix for helping to write the paper.

{\small
\bibliographystyle{ieee}
\bibliography{egbib}
}

\end{document}